\documentclass[lettersize,journal]{IEEEtran}

\usepackage[T1]{fontenc}
\usepackage[utf8]{inputenc}
\usepackage{mathtools}

\usepackage{amssymb,mathrsfs}
\usepackage{amsthm}
\usepackage{bm}
\usepackage{scalerel}
\usepackage{nicefrac}
\usepackage{microtype} 
\usepackage[shortlabels]{enumitem}
\usepackage{graphicx}
\usepackage{epstopdf}
\DeclareGraphicsExtensions{.eps,.png,.jpg,.pdf}

\usepackage{url}
\usepackage{colortbl}
\usepackage{booktabs}
\usepackage{multirow}
\usepackage[normalem]{ulem}
\usepackage{xparse}
\usepackage{calc}
\usepackage{etoolbox}

\makeatletter
\@ifpackageloaded{natbib}{
	\relax
}{
	\usepackage{cite}
}
\makeatother

\usepackage{array}
\newcolumntype{L}[1]{>{\raggedright\let\newline\\\arraybackslash\hspace{0pt}}m{#1}}
\newcolumntype{C}[1]{>{\centering\let\newline\\\arraybackslash\hspace{0pt}}m{#1}}
\newcolumntype{R}[1]{>{\raggedleft\let\newline\\\arraybackslash\hspace{0pt}}m{#1}}

\makeatletter
\let\MYcaption\@makecaption
\makeatother
\makeatletter
\let\@makecaption\MYcaption
\makeatother

\usepackage{glossaries}
\makeatletter
\sfcode`\.1006

\let\oldgls\gls
\let\oldglspl\glspl

\newcommand\fussy@ifnextchar[3]{%
	\let\reserved@d=#1%
	\def\reserved@a{#2}%
	\def\reserved@b{#3}%
	\futurelet\@let@token\fussy@ifnch}
\def\fussy@ifnch{%
	\ifx\@let@token\reserved@d
		\let\reserved@c\reserved@a
	\else
		\let\reserved@c\reserved@b
	\fi
	\reserved@c}

\renewcommand{\gls}[1]{%
\oldgls{#1}\fussy@ifnextchar.{\@checkperiod}{\@}}
\renewcommand{\glspl}[1]{%
\oldglspl{#1}\fussy@ifnextchar.{\@checkperiod}{\@}}

\newcommand{\@checkperiod}[1]{%
	\ifnum\sfcode`\.=\spacefactor\else#1\fi
}

\robustify{\gls}
\robustify{\glspl}
\makeatother

\newacronym{wrt}{w.r.t.}{with respect to}
\newacronym{RHS}{R.H.S.}{right-hand side}
\newacronym{LHS}{L.H.S.}{left-hand side}
\newacronym{iid}{i.i.d.}{independent and identically distributed}

\ifx\notloadhyperref\undefined
	\ifx\loadbibentry\undefined
		\usepackage[hidelinks,hypertexnames=false]{hyperref}
	\else
		\usepackage{bibentry}
		\makeatletter\let\saved@bibitem\@bibitem\makeatother
		\usepackage[hidelinks,hypertexnames=false]{hyperref}
		\makeatletter\let\@bibitem\saved@bibitem\makeatother
	\fi
\else
	\ifx\loadbibentry\undefined
		\relax
	\else
		\usepackage{bibentry}
	\fi
\fi

\usepackage[capitalize]{cleveref}
\crefname{equation}{}{}
\Crefname{equation}{}{}
\crefname{claim}{claim}{claims}
\crefname{step}{step}{steps}
\crefname{line}{line}{lines}
\crefname{condition}{condition}{conditions}
\crefname{dmath}{}{}
\crefname{dseries}{}{}
\crefname{dgroup}{}{}

\crefname{Problem}{Problem}{Problems}
\crefformat{Problem}{Problem~(#2#1#3)}
\crefrangeformat{Problem}{Problems~(#3#1#4) to~(#5#2#6)}

\crefname{Theorem}{Theorem}{Theorems}
\crefname{Corollary}{Corollary}{Corollaries}
\crefname{Proposition}{Proposition}{Propositions}
\crefname{Lemma}{Lemma}{Lemmas}
\crefname{Definition}{Definition}{Definitions}
\crefname{Example}{Example}{Examples}
\crefname{Assumption}{Assumption}{Assumptions}
\crefname{Remark}{Remark}{Remarks}
\crefname{Rem}{Remark}{Remarks}
\crefname{remarks}{Remarks}{Remarks}
\crefname{Appendix}{Appendix}{Appendices}
\crefname{Supplement}{Supplement}{Supplements}
\crefname{Exercise}{Exercise}{Exercises}
\crefname{Theorem_A}{Theorem}{Theorems}
\crefname{Corollary_A}{Corollary}{Corollaries}
\crefname{Proposition_A}{Proposition}{Propositions}
\crefname{Lemma_A}{Lemma}{Lemmas}
\crefname{Definition_A}{Definition}{Definitions}

\usepackage{crossreftools}

\ifx\loadbreqn\undefined
	\relax
\else
	\usepackage{breqn}
\fi

\makeatletter
\def\cleartheorem#1{%
    \expandafter\let\csname#1\endcsname\relax
    \expandafter\let\csname c@#1\endcsname\relax
}
\def\clearthms#1{ \@for\tname:=#1\do{\cleartheorem\tname} }
\makeatother

\ifx\renewtheorem\undefined
	\ifx\useTheoremCounter\undefined
		\newtheorem{Theorem}{Theorem}
		\newtheorem{Corollary}{Corollary}
		\newtheorem{Proposition}{Proposition}
		
	\else
		\newtheorem{Theorem}{Theorem}

	\fi

\fi

\theoremstyle{remark}

\theoremstyle{plain}

\newcommand{\qednew}{\nobreak \ifvmode \relax \else
		\ifdim\lastskip<1.5em \hskip-\lastskip
			\hskip1.5em plus0em minus0.5em \fi \nobreak
		\vrule height0.75em width0.5em depth0.25em\fi}



\NewDocumentCommand{\movedownsub}{e{^_}}{%
	\IfNoValueTF{#1}{%
		\IfNoValueF{#2}{^{}}
	}{%
		^{#1}
	}%
	\IfNoValueF{#2}{_{#2}}
}

\let\latexchi\chi
\RenewDocumentCommand{\chi}{}{\latexchi\movedownsub}

\newcommand{\Real}{\mathbb{R}}

\newcommand{\calE}{\mathcal{E}}

\newcommand{\calG}{\mathcal{G}}

\newcommand{\calQ}{\mathcal{Q}}

\newcommand{\calV}{\mathcal{V}}

\newcommand{\ba}{\mathbf{a}}

\newcommand{\bc}{\mathbf{c}}

\newcommand{\bd}{\mathbf{d}}

\newcommand{\be}{\mathbf{e}}

\newcommand{\bi}{\mathbf{i}}

\newcommand{\bl}{\mathbf{l}}

\newcommand{\bn}{\mathbf{n}}

\newcommand{\bp}{\mathbf{p}}

\newcommand{\bq}{\mathbf{q}}

\newcommand{\br}{\mathbf{r}}

\newcommand{\bs}{\mathbf{s}}

\newcommand{\bt}{\mathbf{t}}


\newcommand{\ifbcdot}[1]{\ifblank{#1}{\cdot}{#1}}

\DeclarePairedDelimiterX\abs[1]{\lvert}{\rvert}{\ifbcdot{#1}}
\DeclarePairedDelimiterX\parens[1]{(}{)}{\ifbcdot{#1}}
\DeclarePairedDelimiterX\brk[1]{[}{]}{\ifbcdot{#1}}
\DeclarePairedDelimiterX\braces[1]{\{}{\}}{\ifbcdot{#1}}
\DeclarePairedDelimiterX\angles[1]{\langle}{\rangle}{\ifblank{#1}{\cdot,\cdot}{#1}}
\DeclarePairedDelimiterX\ip[2]{\langle}{\rangle}{\ifbcdot{#1},\ifbcdot{#2}}
\DeclarePairedDelimiterX\norm[1]{\lVert}{\rVert}{\ifbcdot{#1}}
\DeclarePairedDelimiterX\ceil[1]{\lceil}{\rceil}{\ifbcdot{#1}}
\DeclarePairedDelimiterX\floor[1]{\lfloor}{\rfloor}{\ifbcdot{#1}}

\DeclareFontFamily{U}{matha}{\hyphenchar\font45}
\DeclareFontShape{U}{matha}{m}{n}{
      <5> <6> <7> <8> <9> <10> gen * matha
      <10.95> matha10 <12> <14.4> <17.28> <20.74> <24.88> matha12
      }{}
\DeclareSymbolFont{matha}{U}{matha}{m}{n}
\DeclareFontSubstitution{U}{matha}{m}{n}

\DeclareFontFamily{U}{mathx}{\hyphenchar\font45}
\DeclareFontShape{U}{mathx}{m}{n}{
      <5> <6> <7> <8> <9> <10>
      <10.95> <12> <14.4> <17.28> <20.74> <24.88>
      mathx10
      }{}
\DeclareSymbolFont{mathx}{U}{mathx}{m}{n}
\DeclareFontSubstitution{U}{mathx}{m}{n}

\DeclareMathDelimiter{\vvvert}{0}{matha}{"7E}{mathx}{"17}
\DeclarePairedDelimiterX\vertiii[1]{\vvvert}{\vvvert}{\ifbcdot{#1}}

\DeclarePairedDelimiterXPP\trace[1]{\operatorname{Tr}}{(}{)}{}{\ifbcdot{#1}} 
\DeclarePairedDelimiterXPP\col[1]{\operatorname{col}}{\{}{\}}{}{\ifbcdot{#1}} 
\DeclarePairedDelimiterXPP\row[1]{\operatorname{row}}{\{}{\}}{}{\ifbcdot{#1}} 
\DeclarePairedDelimiterXPP\erf[1]{\operatorname{erf}}{(}{)}{}{\ifbcdot{#1}}
\DeclarePairedDelimiterXPP\erfc[1]{\operatorname{erfc}}{(}{)}{}{\ifbcdot{#1}}
\DeclarePairedDelimiterXPP\KLD[2]{D}{(}{)}{}{\ifbcdot{#1}\, \delimsize\|\, \ifbcdot{#2}} 
\DeclarePairedDelimiterXPP\op[2]{\operatorname{#1}}{(}{)}{}{#2} 

\newcommand{\eqa}[1]{\stackrel{#1}{=}}
\newcommand{\ed}{\eqa{\mathrm{d}}}

\DeclarePairedDelimiterXPP\indicate[1]{{\bf 1}}{\{}{\}}{}{\ifbcdot{#1}}

\providecommand\given{}

\DeclarePairedDelimiterX\Set[2]\{\}{%
\renewcommand\given{\SetSymbol[\delimsize]{#1}}
#2
}
\DeclarePairedDelimiterX\Setc[1]\{\}{%
\renewcommand\given{\SetSymbol{:}}
#1
}

\NewDocumentCommand\set{s o m}{%
	\IfBooleanTF#1%
	{\IfValueTF{#2}{\Set*{#2}{#3}}{\Setc*{#3}}}%
	{\IfValueTF{#2}{\Set{#2}{#3}}{\Setc{#3}}}%
}

\NewDocumentCommand{\evalat}{ s O{\big} m e{_^} }{%
\IfBooleanTF{#1}%
{\left. #3 \right|}{#3#2|}%
\IfValueT{#4}{_{#4}}%
\IfValueT{#5}{^{#5}}%
}

\providecommand\given{}
\DeclarePairedDelimiterXPP\cprob[1]{}(){}{
\renewcommand\given{\nonscript\,\delimsize\vert\allowbreak\nonscript\,\mathopen{}}%
#1%
}
\DeclarePairedDelimiterXPP\cexp[1]{}[]{}{
\renewcommand\given{\nonscript\,\delimsize\vert\allowbreak\nonscript\,\mathopen{}}%
#1%
}

\DeclareDocumentCommand \P { s e{_^} d() g } {%
	\mathbb{P}%
	\IfBooleanTF{#1}%
		{
			\IfValueT{#2}{_{#2}}%
			\IfValueT{#3}{^{#3}}%
			\IfValueTF{#5}{\cprob{#4 \given #5}}{\IfValueT{#4}{\cprob{#4}}}%
		}%
		{
			\IfValueT{#2}{_{#2}}%
			\IfValueT{#3}{^{#3}}%
			\IfValueTF{#5}{\cprob*{#4 \given #5}}{\IfValueT{#4}{\cprob*{#4}}}%
		}%
}

\DeclareDocumentCommand \E { s e{_^} o g } {%
	\mathbb{E}%
	\IfBooleanTF{#1}%
		{
			\IfValueT{#2}{_{#2}}%
			\IfValueT{#3}{^{#3}}%
			\IfValueTF{#5}{\cexp{#4 \given #5}}{\IfValueT{#4}{\cexp{#4}}}%
		}%
		{
			\IfValueT{#2}{_{#2}}%
			\IfValueT{#3}{^{#3}}%
			\IfValueTF{#5}{\cexp*{#4 \given #5}}{\IfValueT{#4}{\cexp*{#4}}}%
		}%
}

\ExplSyntaxOn
\NewDocumentCommand \dist {m o o} {%
\mathrm{#1}\left(%
	\IfValueT{#3}{%
		\tl_if_blank:nTF{ #3 }{\cdot\, \middle|\, }{#3\, \middle|\, }%
	}
	\IfValueT{#2}{#2}%
\right)%
}
\ExplSyntaxOff

\NewDocumentCommand {\cbrace} {t+ D[]{black} D(){\widthof{#5}} m m } {%
	\begingroup%
		\color{#2}
		\IfBooleanTF{#1}{%
			\overbrace{#4}^%
		}{
			\underbrace{#4}_%
		}%
		{\parbox[c]{#3}{\centering\footnotesize{#5}}}%
	\endgroup%
}

\let\oldforall\forall
\renewcommand{\forall}{\oldforall \, }

\let\oldexist\exists
\renewcommand{\exists}{\oldexist \, }

\graphicspath{{./Figures/}{./figures/}}
\DeclareDocumentCommand{\includeCroppedPdf}{ o O{./Figures/} m }{
	\IfFileExists{#2#3-crop.pdf}{}{%
		\immediate\write18{pdfcrop #2#3.pdf #2#3-crop.pdf}}%
	\includegraphics[#1]{#2#3-crop.pdf}
}

\makeatletter
\newcommand*{\addFileDependency}[1]{
  \typeout{(#1)}
  \@addtofilelist{#1}
  \IfFileExists{#1}{}{\typeout{No file #1.}}
}
\makeatother

\definecolor{gray90}{gray}{0.9}

\ifx\nohighlights\undefined
	\newcommand{\red}[1]{{\color{red} #1}}
	\newcommand{\blue}[1]{{{\color{blue} #1}}}

	\newcommand{\msout}[1]{\text{\color{green} \sout{\ensuremath{#1}}}}
	
	\newcommand{\del}[1]{{\color{green}\ifmmode \msout{#1}\else\sout{#1}\fi}}
\else
        
\else
	\newcommand{\red}[1]{#1}
	\newcommand{\blue}[1]{#1}

	\newcommand{\msout}[1]{#1}
	\newcommand{\del}[1]{#1}
\fi

\newcommand{\hhide}[1]{}

\newcommand{\txp}[2]{\texorpdfstring{#1}{#2}}

\ifx\diagnoselabel\undefined
	\relax
\else
	\makeatletter
	\def\@testdef #1#2#3{%
		\def\reserved@a{#3}\expandafter \ifx \csname #1@#2\endcsname
			\reserved@a  \else
			\typeout{^^Jlabel #2 changed:^^J%
				\meaning\reserved@a^^J%
				\expandafter\meaning\csname #1@#2\endcsname^^J}%
			\@tempswatrue \fi}
	\makeatother
\fi

\newcommand{\bit}[1]{\textbf{\textit{#1}}}
\newcommand{\tb}[1]{\textbf{#1}}

\def\p{\partial}

\DeclareFontFamily{U}{MnSymbolC}{}
\DeclareSymbolFont{MnSyC}{U}{MnSymbolC}{m}{n}
\DeclareMathSymbol{\diamondplus}{\mathbin}{MnSyC}{"7C}
\DeclareMathSymbol{\diamonddot}{\mathbin}{MnSyC}{"7E}
\DeclareFontShape{U}{MnSymbolC}{m}{n}{
    <-6>  MnSymbolC5
   <6-7>  MnSymbolC6
   <7-8>  MnSymbolC7
   <8-9>  MnSymbolC8
   <9-10> MnSymbolC9
  <10-12> MnSymbolC10
  <12->   MnSymbolC12}{}

\def\p{\partial}

\usepackage[caption=false,font=normalsize,labelfont=sf,textfont=sf]{subfig}
\usepackage{textcomp}
\usepackage{stfloats}

\usepackage{algorithm}
\usepackage{algorithmic}

\DeclareMathOperator{\st}{s.t.\ }

\def\bcon{\begin{convention}} 
\def\econ{\end{convention}}
\def\bc{\begin{cora}}
\def\ec{\end{cora}}
\def\bcl{\begin{rema}}
\def\ecl{\end{rema}}
\def\bd{\begin{defa}}
\def\ed{\end{defa}}
\def\ben{\begin{enumerate}}
\def\benr{\begin{enumerate}[label=(\roman*)]}
\def\een{\end{enumerate}}
\def\be{\begin{align}}
\def\ee{\end{align}}
\def\bse{\begin{align*}}
\def\ese{\end{align*}}
\def\bex{\begin{exma}}
\def\eex{\end{exma}}
\def\bexe{\begin{exercise}}
\def\eexe{\end{exercise}}
\def\bit{\begin{itemize}}
\def\eit{\end{itemize}}
\def\bl{\begin{lema}}
\def\el{\end{lema}}
\def\bnn{\begin{rema}}
\def\enn{\end{rema}}
\def\bn{\begin{note}}
\def\en{\end{note}}
\def\bpo{\begin{postulate}}
\def\epo{\end{postulate}}

\def\br{\begin{rema}}
\def\er{\end{rema}}
\def\bs{\begin{solution}}
\def\es{\end{solution}}
\def\btab{\begin{table}}
\def\etab{\end{table}}
\def\btb{\begin{tabular}}
\def\etb{\end{tabular}}
\def\bter{\begin{defa}}
\def\eter{\end{defa}}
\def\bt{\begin{thma}}
\def\et{\end{thma}}
\def\btik{\begin{tikzpicture}}
\def\etik{\end{tikzpicture}}
  \def\bi{\begin{IEEEeqnarray*}}
  \def\ei{\end{IEEEeqnarray*}}
    \def\ba{\begin{array}}
  \def\ea{\end{array}}

\def\b{\beta}
\def\del{\delta}

\def\d{\mathrm{d}}

\def\b1{\mathbb{1}} 

\def\R{\mathbb{R}} 

\def\bbox{\begin{tcolorbox}[colback=white!5,colframe=blue!75!black,title=Exercise]}
\def\ebox{\end{tcolorbox}}

\newcommand{\cl}{:\text{ }}

 \def\bse{\begin{equation*}}
  \def\ese{\end{equation*}}
   \def\btab{\begin{table}}
  \def\etab{\end{table}}
  \def\btb{\begin{tabular}}
  \def\etb{\end{tabular}}
\usepackage{tikz}
\usepackage{tikz,pgfplots}

\usepackage{tikz-cd}
\newcommand{\first}[1]{\red{\tb{#1}}}
\newcommand{\second}[1]{\blue{\tb{#1}}}

\newcommand{\third}[1]{{#1}}

\newcommand{\exc}[1]{{#1}}

\begin{document}

\title{Node Embedding from Hamiltonian Information Propagation in Graph Neural Networks}

\author{Qiyu Kang$^{1*}$, Kai Zhao$^{1*}$, Yang~Song$^{2}$, Sijie~Wang$^{1}$, Rui~She$^{1}$, and Wee Peng Tay$^{1}$,  \IEEEmembership{Senior Member,~IEEE}
\thanks{$^{*}$ Equal contribution.}
\thanks{$^{1}$The authors are with the School of Electrical and Electronic Engineering, Nanyang Technological University, Singapore. Email: \{kang0080@e., kai.zhao@, wang1679@e., rui.she@, and wptay@\}ntu.edu.sg.} 
\thanks{$^{2}$The author is with the C3 AI, Singapore. Email:  yang.song@c3.ai.} 
}

\markboth{IEEE Transactions on Neural Networks and Learning Systems}%
{Shell \MakeLowercase{\textit{et al.}}: A Sample Article Using IEEEtran.cls for IEEE Journals}


\maketitle

\begin{abstract}
Graph neural networks (GNNs) have achieved success in various inference tasks on graph-structured data. However, common challenges faced by many GNNs in the literature include the problem of graph node embedding under various geometries and the over-smoothing problem. 
To address these issues, we propose a novel graph information propagation strategy called Hamiltonian Dynamic GNN (HDG) that uses a Hamiltonian mechanics approach to learn node embeddings in a graph. The Hamiltonian energy function in HDG is learnable and can adapt to the underlying geometry of any given graph dataset. We demonstrate the ability of HDG to automatically learn the underlying geometry of graph datasets, even those with complex and mixed geometries, through comprehensive evaluations against state-of-the-art baselines on various downstream tasks. We also verify that HDG is stable against small perturbations and can mitigate the over-smoothing problem when stacking many layers.
\end{abstract}


\section{Introduction}


\IEEEPARstart{M}{any} real-world objects and datasets, such as social media networks, molecular graphs in chemistry, and citation networks, are characterized by graph structures.
Analyzing information from these graphs has become increasingly important. 
To address this challenge, various graph neural networks (GNNs) \cite{yueBio2019, AshoorNC2020, kipf2017semi, ZhangTKDE2022, WuTNNLS2021} have been proposed. Most GNNs adopt the message-passing paradigm, which involves using learnable non-linear functions to propagate information across nodes in the graph. Despite their effectiveness, these GNNs suffer from the well-known over-smoothing phenomenon, which is caused by repeated message passing that can blur the feature representations of nodes and result in a loss of discriminative power. Furthermore, many existing GNNs embed graph nodes into Euclidean spaces, even when the inherent graph structure is highly complex. Euclidean embeddings may not always be able to capture the complex relationships between nodes in the graph. For instance, in a complete binary tree with exponentially increasing nodes at increasing depths, the distances between leaf nodes may not be faithfully represented by Euclidean embeddings. To address this issue, some researchers \cite{liu2019hyperbolic_GNN,gu2018learning,bachmann2020constant_curvature,lou2020differentiating} use (products of) constant curvature Riemannian spaces for graph node embedding where the spaces are assumed to be spherical, hyperbolic, or Euclidean. 

Recently, many attempts have been made to either learn physical laws by neural networks \cite{greydanus2019hamiltonian, zhongiclr2020, ChenNeurIPS2021} or model neural networks using physical laws \cite{haber2017stable,huang22a,duong21hamiltonian}.
For instance, interaction networks and graph networks have been developed to simulate increasingly complex physical systems, allowing researchers to model and understand the behavior of particles and other physical entities in a wide range of contexts \cite{battaglia2016interaction,sanchez2018graph,lilearning}. Recently, neural ordinary differential equations (ODEs) \cite{chen2018neural} have been proposed and  used to stabilize the neural network training process \cite{haber2017stable}, improve the adversarial robustness \cite{huang22a, yan2019robustness, kang2021Neurips}, and control the autonomous operation of mobile robots \cite{duong21hamiltonian,WanKanSheC23}. 
The dynamics of a physical system under basic physical laws can often be described by differential equations \cite{raissi2017physics}, such as the heat diffusion equation or the Hamiltonian equations used in classical mechanics for the $n$-body astronomical systems \cite{raissi2017physics, constantin2020navier, lee2013smooth, greengard1990numerical, march1995many}. 
Researchers have been inspired by physics to model the way nodes exchange features in a graph by using diffusion equations, which can be learned through neural Partial Differential Equations (PDEs) \cite{chamberlain2021grand,chamberlain2021blend,SonKanWan:C22}. For instance, \cite{chamberlain2021grand} models the message passing process, i.e., features exchanging between nodes, as the heat diffusion, while \cite{chamberlain2021blend, SonKanWan:C22} use Beltrami diffusion. These PDE-based GNNs have been shown to achieve competitive performance compared to conventional GNNs on various graph-based learning tasks. In addition, some works like \cite{rusch2022icml} have modeled the graph nodes as coupled oscillators with a coupled oscillating ODE guiding the message passing process. This graph-coupled oscillator network mitigates the over-smoothing problem that is commonly encountered in GNNs.

In this work, we incorporate physics-inspired inductive biases into new GNN architectures.
The core idea is that a GNN enables the transfer of information among nodes in a graph through edges, much like the way energy moves between bodies in an $n$-body system via gravitational field. Rather than relying on physical laws to govern the system's dynamics, we interpret a graph as an $n$-body system that evolves based on a learnable law derived from learnable Hamiltonian vector fields and their associated differential equations. This learnable law, which we refer to as the \tb{learnable node embedding law}, guides the evolution of features and embeds them into a manifold to support downstream tasks such as node classification or link prediction.

In contrast to the aforementioned approaches in the literature, the new proposed GNN paradigm is based on Hamiltonian mechanics \cite{Goldstein2001} to utilize its energy-conservative nature. 
The model dynamics are guided by Hamiltonian mechanics \cite{greengard1990numerical}, with the message passing being implicitly guided by the learnable Hamiltonian energy function. We call our model the \emph{Hamiltonian Dynamic GNN (HDG)}.
HDG offers several advantages over existing GNNs. Firstly, it effectively addresses the over-smoothing problem commonly encountered by many GNNs in the literature. This is achieved through the use of a learnable node embedding law that guides the evolution of node features while ensuring that they remain constrained to the Hamiltonian orbit, following the law of conservation of energy \cite{da2008lectures} (cf.\ \cref{thm:constantE}). Secondly, HDG generalizes the geodesic tools used in Riemannian graph node embedding \cite{liu2019hyperbolic_GNN,chami2019hyperbolicGCNN,gu2018learning,bachmann2020constant_curvature,lou2020differentiating,xiong2022pseudo} by using the Hamiltonian orbit formulation (cf.\ equation \cref{eq:gpq}). Unlike existing works that assume fixed manifolds with strict constraints, such as product spaces of constant curvatures, HDG allows node embedding in a manifold with flexible local geometry. This feature provides a more accurate and flexible representation of the underlying graph structure, making HDG suitable for a wider range of graph-structured data.

\tb{Main contributions.} Our main contributions are summarized as follows:
\begin{enumerate}
    \item Inspired by the $n$-body physical system, we present a new GNN information propagation mechanism, which is able to embed node features following Hamiltonian flows under the influence of neighbors.
    \item By the conservative nature of Hamilton equations, HDG enables a stable inference process that is robust against over-smoothing while updating the node features over time or layers.
    \item The $n$-body system is known to be chaotic \emph{under basic physical laws}. In contrast, we show that HDG \emph{with a learnable node embedding law} is a stable system under small input perturbations. 
    \item As a generalization of Riemannian graph node embedding, HDG automatically learns the underlying geometry of any given graph dataset, even if it has diverse geometries, without requiring extensive tuning. We demonstrate this ability empirically by evaluating HDG on two downstream tasks: node classification and link prediction.
\end{enumerate}
The rest of this paper is organized as follows. Related work is discussed in \cref{sec:rel_wor}.  In \cref{sec:pre}, we present a brief overview of some basic concepts related to the $n$-body system, Hamiltonian mechanics, and Riemannian graph node embedding. Our model and framework are presented in \cref{sec:pro_fra}. We present extensive experimental results in \cref{sec:exp}. We conclude our paper in \cref{sec:con}.

\begin{figure*}
\centering
  \includegraphics[width=0.6\textwidth]{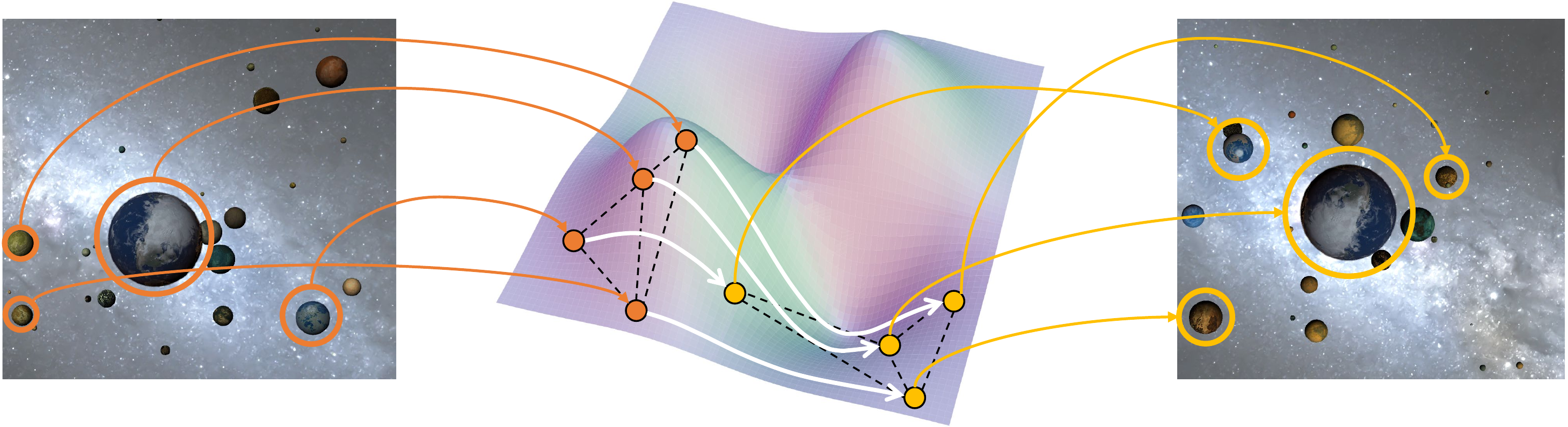}
  \caption{The graph node features evolve on a manifold following a \emph{learnable} law analogous to how the $n$-body system evolves in $\Real^3$ space following basic physical laws. Each node in the graph is analogous to a body in the  $n$-body system, while the edges indicate the ``forces'' (i.e., relations) between nodes.}
  \label{fig:teaser}
\end{figure*}

\section{Related Work}\label{sec:rel_wor}
In this work, we introduce a novel Hamiltonian graph neural network. This is distinct from previous methods that use Hamiltonian-inspired neural networks without a graph structure or to simulate physical Hamiltonian systems. In the following sections, we briefly review existing Hamiltonian neural networks and explore the intersection between physics and GNNs. We also examine Hamiltonian equations from a node embedding perspective, which can be seen as exponential maps that embed nodes into a flexible manifold via its symplectic cotangent bundle. This connection allows us to draw parallels with Riemannian manifold GNNs, which embed nodes into specialized Riemannian manifolds using their associated exponential maps.

\subsection*{Solving the $n$-Body Problem Using DNNs}
The $n$-body system is a well-known example of a chaotic system, which means that even in the absence of randomness, a tiny change in the initial state can result in significantly different evolutionary trajectories \cite{strogatz2018nonlinear}. 
To address the practical difficulties of solving the equations of motion for an $n$-body system, \cite{Breen_2020} proposes to solve a three-body problem using DNNs. The results suggest that a trained DNN can replace existing numerical solvers to enable fast and scalable simulations of $n$-body systems, providing accurate enough predictions of periodic orbits with arbitrary masses for physicists and astronomers \cite{LIAO2022101850}. Similarly, in \cite{chen2019symplectic}, the authors propose to learn a Hamiltonian function of the system by a neural network to capture the dynamics of physical systems from observed trajectories. They apply the system to the three-body system and show good prediction performance. 
To forecast dynamics, the work \cite{choudhary2020physics} use neural networks that incorporate Hamiltonian dynamics to efficiently learn phase space orbits even as nonlinear systems transition from order to chaos. The authors demonstrate the effectiveness of Hamiltonian neural networks on several dynamics benchmarks.

\emph{In this paper, we are neither simulating an $n$-body system nor solving a real-world $n$-body physics problem. Instead, we combine the $n$-body concept with graph neural networks to develop a new graph node manifold embedding strategy. }
In the physics $n$-body system, all bodies interact with each other directly through the gravitational field, and the system is known to be a chaotic system in general.
In contrast, HDG only allows for interaction between a node and its neighborhoods at each time, and the learnable node embedding law is learned through gradient-based back-propagation for the node classification task instead of regressing basic physical laws. In \cref{fig:converge}, we visualize the feature evolution in HDG at different time steps and observe steady feature evolution instead of chaos.

\subsection*{Hamiltonian Neural Networks}
In the literature, researchers \cite{haber2017stable,huang22a}  have applied Hamiltonian equations to train neural networks by conserving an energy-like quantity using Hamiltonian dynamics. To infer the dynamics of a physical system using data-driven approaches, \cite{greydanus2019hamiltonian, zhongiclr2020, ChenNeurIPS2021} have designed Hamiltonian neural network systems and used neural ODE solvers \cite{chen2018neural} to update the system. The work  \cite{haber2017stable} builds Hamiltonian-inspired neural ODE to stabilize the gradients so as to avoid vanishing and exploding gradients during network training. Compared to traditional neural networks, Hamiltonian neural networks have several advantages such as the ability to preserve energy and the natural incorporation of conservation laws. They also provide a more interpretable and physically meaningful representation of the system.


\emph{
Different from the above Hamiltonian neural networks, our paper proposes a novel graph information propagation mechanism using a Hamiltonian mechanics approach to learn node embeddings in a graph. Our approach adapts a learnable Hamiltonian energy function to the underlying geometry of any given graph dataset and addresses challenges faced by GNNs in graph node embedding and the over-smoothing problem. 
}

\subsection*{Graph Neural Diffusion}
Neural Partial Differential Equations (PDEs) have been applied to graph-structured data \cite{chamberlain2021grand, chamberlain2021blend,SonKanWan:C22}, where different diffusion schemes are assumed when performing message passing on graphs. To be more specific, the heat diffusion model is assumed in \cite{chamberlain2021grand} and the Beltrami diffusion model is used in \cite{chamberlain2021blend,SonKanWan:C22}. The paper \cite{rusch2022icml} models the nodes in a graph as coupled oscillators and message passing can be performed using coupled oscillators, i.e., a second-order ODE. 

\emph{In this paper, we assume that the evolution of node features satisfies the Hamiltonian equations, analogous to the $n$-body system, which is different from the above graph neural diffusion mechanics.}

\subsection*{Riemannian Manifold GNNs}
The majority of GNNs in the literature embed graph-structured data into Euclidean spaces, as seen in works such as
\cite{yueBio2019,AshoorNC2020,kipf2017semi,ZhangTKDE2022,WuTNNLS2021}. 
They perform well on datasets with high Gromov $\delta$-hyperbolicity distributions (i.e., datasets that are less hyperbolic in nature), such as the Cora dataset \cite{McCallum2004AutomatingTC}.
However, for datasets with tree-like graph structures and low $\delta$-hyperbolicities, embedding should more appropriately be in a hyperbolic space. Examples include the Disease \cite{chami2019hyperbolic_GCNN} and Airport \cite{chami2019hyperbolic_GCNN} datasets, on which the node embeddings learned by these GNNs tend to produce poorer inference outcomes.
To better handle such hierarchical graph data, \cite{liu2019hyperbolic_GNN,chami2019hyperbolic_GCNN,zhang2021lorentzian,zhu2020GIL} proposed to embed nodes into a hyperbolic space, thus yielding hyperbolic GNNs. 
Moreover, \cite{zhu2020GIL} proposes a mixture of embeddings from Euclidean and hyperbolic spaces. This mixing operation relaxes the strong space assumption of using only one type of space for a dataset. 
In recent studies, like \cite{gu2018learning,bachmann2020constant_curvature,lou2020differentiating}, researchers employ (products of) Riemannian spaces with constant curvature for embedding graph nodes. These spaces are assumed to be either spherical, hyperbolic, or Euclidean. The study by \cite{xiong2022pseudo} explores a special type of pseudo-Riemannian manifold known as the pseudo-hyperboloid for graph node embedding. The pseudo-hyperboloid possesses constant non-zero curvature and is diffeomorphic to the combination of a unit sphere and Euclidean space.

\emph{Embedding nodes in the aforementioned \emph{restricted} Riemannian manifolds is achieved through the exponential map in closed forms, which is essentially a geodesic curve on the manifolds as the projected curve of the \emph{cogeodesic orbits} on the manifolds' cotangent bundles \cite{lee2013smooth,klingenberg2011riemannian}. In our work, we propose to generalize this approach by embedding nodes into a more general manifold using Hamiltonian orbits, allowing our model to adapt flexibly to the inherent geometry of graph datasets.
}

\emph{Notations:} We use the \emph{Einstein summation convention} \cite{lee2013smooth} for expressions with indices: if in any term, the same index appears twice as both an upper and a lower index, that term is assumed to be summed over all possible values of that index. 

\section{Preliminaries}\label{sec:pre}
In this section, we provide a brief overview of the $n$-body system model \cite{lee2013smooth,greengard1990numerical} in Hamiltonian mechanics, which describes the evolution of particles in $\Real^3$ space governed by physical laws. We then establish a connection between the exponential maps used in the Riemannian manifold GNN literature \cite{chami2019hyperbolic_GCNN,gu2018learning,bachmann2020constant_curvature,lou2020differentiating} and Hamiltonian mechanics. 
In \cref{sec:pro_fra}, we apply concepts introduced in this section to GNNs, treating a graph as a Hamiltonian system that evolves based on a learnable node embedding law.

The $n$-body system in physics models the gravitational interaction between $n$ astronomical objects. More formally,
consider $n$ astronomical objects moving in $\mathbb{R}^3$ space, and suppose their positive masses are $m_1, \ldots, m_n$, respectively. 
We denote the position of the $k$-th object in $\Real^3$ at time $t$ as $\left(q_k^1(t), q_k^2(t), q_k^3(t)\right)$. The evolution of the $n$-body system over time is a curve in a ${3n}$-dimensional manifold $Q$ given by
\begin{align}
q(t)=\left(q_1^1(t), q_1^2(t), q_1^3(t), \ldots, q_n^1(t), q_n^2(t), q_n^3(t)\right).  \label{eq:nbod_allobjs}
\end{align}
The net force $\left(F_k^1(q), F_k^2(q), F_k^3(q)\right)$ on the $k$-th object depends only on the positions $q(t)$ of all the objects via gravitational forces. Newton's second law of motion states that the objects will move along the trajectories satisfying the following second-order ODE:
\begin{align}
\begin{aligned}
&m_k \ddot{q}_k^1(t)=F_k^1(q(t)), \\
&m_k \ddot{q}_k^2(t)=F_k^2(q(t)), \\ \label{eq:nbody_0}
&m_k \ddot{q}_k^3(t)=F_k^3(q(t)), \quad k=1, \ldots, n .
\end{aligned}
\end{align}
where we use $\dot{q}$ and $\ddot{q}$ to denote the first- and second-order ordinary derivatives of $q(t)$, respectively. 
In physics, we could further define 
\begin{align}\label{eq:nbod_mom}
    \left(p_k^1(t), p_k^2(t), p_k^3(t)\right)\coloneqq \left(m_k\dot{q}_k^1(t), m_k\dot{q}_k^2(t), m_k \dot{q}_k^3(t)\right),
\end{align} which is the \tb{momentum vector} of the $k$-th object at time $t$. The momentum vector specifies each body's evolutional velocity in $\Real^3$.

If we further relabel the index and write $$q(t)=(q^1(t),\ldots, q^{3n}(t))$$ and $$F(q)=\left(F_1,\ldots, F_{3 n}(q)\right),$$ we have a more compact representation of \cref{eq:nbody_0} as follows:
\begin{align}\label{eq:momentum}
M_{i j} \ddot{q}^j(t)=F_i(q(t)) .
\end{align}
where the matrix $M=\left[M_{i j}\right]_{i,j=1}^{3n}$ denotes the $3 n \times 3 n$ diagonal matrix whose diagonal entries are $\left(m_1, m_1, m_1, m_2, m_2, m_2, \ldots, m_n, m_n, m_n\right)$. Note that the Einstein summation convention is used here, i.e., there is a summation over the index $j$ on the left-hand side of \cref{eq:momentum}. 
We further concatenate the momentum vector in \cref{eq:nbod_mom} and relabel the index to obtain the full momentum vector
\begin{align}\label{eq:mom}
     p(t)=(p^1(t),\ldots, p^{3n}(t)).
\end{align}
It follows that
\begin{align*}
\begin{aligned}
\dot{q}^i(t) &=M^{i j} p_j(t), \\
\dot{p}_i(t) &=F(q(t)),
\end{aligned}
\end{align*}
where $\left[M^{i j}\right]$ is the inverse of the matrix of $\left[M_{i j}\right]$. 

Under the condition that forces are conservative \cite{de2011generalized,lee2013smooth}, there exists a smooth function $V$, called the {potential energy} of the system, such that $F=-\d V$.  Finally, we can  define the \tb{total energy of the system} as $E$ by
\begin{align}\label{eq:total_E}
E(q, p)=V(q)+K(p),
\end{align}
where $K$ is the {kinetic energy}, defined by
\begin{align}
K(p)=\frac{1}{2} M^{i j} p_i p_j .
\end{align}
We conclude that Newton's second law, therefore, ensures that the system evolves according to the trajectory that solves the following \tb{Hamiltonian equations}:
\begin{align}
\begin{aligned}
\dot{q}^i(t) &= \frac{\p E}{\p p^i},\\
\dot{p}_i(t) &= -\frac{\p E}{\p q^i}. \label{eq:nbody}
\end{aligned}
\end{align}

From the Hamiltonian mechanics perspective, \cref{eq:nbody} are Hamilton's equations for the Hamiltonian flow of energy $E$. From \cite{da2008lectures}, we know that the total Hamiltonian energy $E$ is constant along the trajectory of its induced Hamiltonian flow. This is known as \tb{the law of conservation of energy} in physics:

\begin{Theorem}[Energy conservation law \cite{da2008lectures}]\label{thm:constantE}
If the $n$-body system evolves according to \cref{eq:nbody}, the total energy $E(q(t),p(t))$ of the system remains constant. More generally, for any continuous energy function $E$, the same energy conservation property holds.
\end{Theorem}

We next provide additional background on Hamiltonian mechanics from a differential geometric perspective and investigate the relationship between our approach and Riemannian manifold GNNs \cite{gu2018learning,bachmann2020constant_curvature,lou2020differentiating}.

For each point $q$ on a smooth manifold $Q$ \cite{lee2013smooth}, we can associate it with a $d$-dimensional vector space $T^*_qQ$, called the cotangent space at $q$. We define the cotangent bundle of $Q$ as 
    \begin{align}
T^*Q \coloneqq \coprod_{q\in Q} T^*_qQ.
\end{align}
where $\coprod$ is the disjoint union operation. The space $T^*_qQ$ is in fact a $2d$-dimensional manifold equipped with the \emph{canonical projection map}
\begin{align}
\begin{aligned}
    {\pi} \cl  T^*Q & \to  Q\\
 x & \mapsto  \pi(x), \label{appeq:co_pro}
\end{aligned}
\end{align}
where $\pi(x)$ sends each vector in $T^*_qQ$ to the point $q$ at which it is cotangent. 
\begin{figure}
    \centering
    \includegraphics[width=0.3\textwidth]{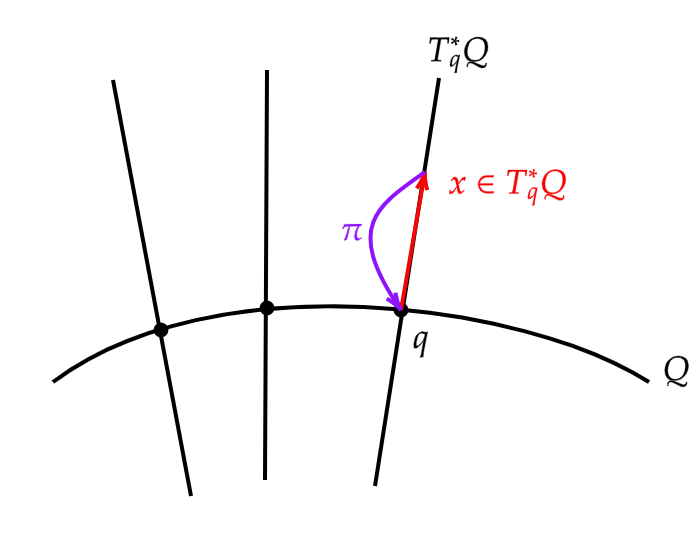}
    \caption{Visualization of a 1-dimensional manifold and its cotangent bundle.}
    \label{fig:my_label}
\end{figure}
In the context of the $n$-body system, we have $d=3n$ as the dimension, and the cotangent bundle $T^*Q$ is a $6n$-dimensional manifold. The points on $T^*Q$ are represented locally as $(q,p)$ with $q\in Q$ and $p$ a cotangent vector in the cotangent space $T^*_qQ$. For a point $q$, we can associate it with different vectors from $T^*_qQ$ which represents different \tb{momenta} for the  $n$-body system. Since we have concatenated all the momentum of each object into $p$ in \cref{eq:mom}, this indicates different momenta for different objects in the  $n$-body system.

For a manifold $Q$, we can equip it with a metric tensor $g$ so that $Q$ now has an additional shape structure. The obtained manifold is called the Riemannian manifold. At each point  $q$, the metric tensor is a $(0,2)$-tensor $g$ that is  symmetric, i.e.,
\begin{align}
g(q)\cl T_qQ &\times T_qQ \to \R \\
\st g(q)(X_1, X_2) &= g(q)(X_2,X_1) \ \forall X_1,X_2\in  T_qQ
\end{align}
where $T_qQ$, the tangent vector space, is the dual vector space of $T^*_qQ$. Intuitively, the metric tensor assigns to the manifold $Q$ a ``shape'' by specifying the norms of tangent vectors and the angles between them. The matrix form of $g$ at each point is a symmetric matrix $[g_{ij}]$. 

The flexibility of $E$ in \cref{eq:nbody} allows for different flow structures on the manifold. For instance, suppose $E$ in \cref{eq:nbody} is defined by
 \begin{align}
        E(q, p)=\frac{1}{2} g^{i j}(q) p_i p_j, \label{eq:gpq}
    \end{align}
with $[g^{ij}]$ denote the inverse matrix of the metric $[g_{ij}]$ defined for $Q$. Then, the canonical projection of the solution $(q,p)$ of \cref{eq:nbody} to $p$ degenerates to the \tb{exponential map} defined on the Riemannian manifold $Q$, starting from one point on the manifold and ending at another point following a metric-induced vector field \cite{lee2006riemannian}. Hyperbolic GNNs \cite{chami2019hyperbolic_GCNN, liu2019hyperbolic_GNN} make use of the exponential map defined on a Riemannian manifold $Q$ with the Minkowski metric. 

To summarize, we have demonstrated that the flexibility of $E$ in \cref{eq:nbody} enables diverse flow structures on the manifold, with various applications in the $n$-body system and Riemannian manifold GNNs:
\begin{itemize}
    \item In the $n$-body system, the energy function $E$ takes the form of \cref{eq:total_E}, and the dynamics of the system are governed by the Hamiltonian equations \cref{eq:nbody} based on basic physical laws.
    \item In the Riemannian manifold GNNs, the energy function $E$ has the form \cref{eq:gpq}, and the Hamiltonian equations \cref{eq:nbody} lead to the exponential map used for embedding graph nodes on manifolds other than the Euclidean space.
\end{itemize}
In HDG, we do not assume the energy $E$ of the system is determined by a fixed function such as \cref{eq:total_E} or \cref{eq:gpq}. Instead, we propose a learnable energy function $E$ for the entire graph, as discussed in the next section. This learnable energy function guides the evolution of graph features according to a learnable node embedding law, which is similar to how the $n$-body system evolves under the influence of basic physical laws.

\section{Proposed Framework}\label{sec:pro_fra}
We consider an undirected graph $\calG = (\calV, \calE)$ consisting of a finite set $\calV$ of vertices, together with a subset $\calE \subset \calV \times \calV$ of edges. 
In GNNs, typically each node $k$ is associated with a $r$-dimensional feature vector $\bq_k=(q_k^1,\ldots,q_k^r)$. 
In general,  GNN architectures facilitate information propagation between nodes in a graph through the edges connecting them, similar to the energy exchange and force propagation between bodies in an $n$-body system through the gravitational field. This analogy inspires us to develop a new type of GNN architecture that propagates information between nodes in a graph using Hamiltonian mechanics. As we show in \cref{sec:pre}, this Hamiltonian mechanics-inspired information propagation also generalizes the Riemannian manifold GNNs.

\subsection{Hamiltonian Information Propagation}
Similar to the evolutionary trajectory of the $n$-body system in the ${3n}$-dimensional manifold $Q$, we take the $|\calV|$ vertices' features as a point in a $r|\calV|$-dimensional manifold $Q_{\calG}$ with index relabeling:
\begin{align}
    q =  \left(q^1,\ldots q^{r|\calV|} \right)=\left( \bq_1,\ldots, \bq_{|\calV|} \right).\label{eq.con_q}
\end{align}

The interaction between graph node features in GNNs are analogous to the gravitational interaction between astronomical objects in the $n$-body system.
In \cref{sec:pre}, the momentum of the $k$-th astronomical object at time $t$ in \cref{eq:nbod_mom} plays an important role. 
Inspired by the  momentum vector, 
we associate to each node $k$ an additional $r$-dimensional \emph{learnable momentum vector} $\bp_k$ which may be regarded as  the ``momentum'' of the node~$k$. This learnable momentum vector guides the direction and speed of its feature update, allowing the node feature to evolve along Hamiltonian orbits on the manifold. More specifically, we set 
\begin{align}
    \bp_k = P_{\mathrm{net}}(\bq_k) \label{eq:mom2}
\end{align}
where $P_{\mathrm{net}}$ is a neural network instantiated by a fully connected (FC) layer.
By concatenating all the ``momentum'' vectors together, we obtain the $r|\calV|$-dimensional vector
\begin{align}
    p = \left(p_1,\ldots p_{r|\calV|} \right)= \left( \bp_1,\ldots, \bp_{|\calV|} \right), \label{eq.con_p}
\end{align}
which can be regarded as a covector in the cotangent bundle of the manifold $Q_{\calG}$. We have  
\begin{align*}
    (q,p) \in T^*Q_{\calG},
\end{align*}
where $T^*Q_{\calG}$ is the cotangent bundle of the manifold $Q_{\calG}$.

In principle, any Hamiltonian dynamical system can be modeled in the form of \cref{eq:nbody} with different system energy functions $E$. 
In a physical $n$-body system, the evolution law is fixed in \cref{eq:total_E} as a total energy formulation that includes both potential and kinetic energy. However, for our node embedding task, it does not make sense to assign energy as potential or kinetic.
In our work, instead of defining an explicit energy function $E$ such as \cref{eq:total_E} or \cref{eq:gpq}, we use a learnable energy function $E_{\mathrm{net}}$ that is parameterized by a neural network introduced in \cref{sec.enet}.  
We let the graph features evolve according to the learnable law, analogous to physical laws. More specifically, we design a new information propagation as the following ODE:
\begin{align}
\begin{aligned}
\dot{q}^i(t) &= \frac{\p E_{\mathrm{net}}}{\p p_i},\\
\dot{p}_i(t) &= -\frac{\p E_{\mathrm{net}}}{\p q^i}, \label{eq:graph_obj}
\end{aligned}
\end{align}
with the initial features at time $t=0$ being the input node features and learnable momentum, $(q(0), p(0)) = (q,p)$.

\subsection{Energy Net \txp{$E_{\mathrm{net}}$}{Enet}} \label{sec.enet}
In the $n$-body system, each object interacts with all the other objects via gravitational forces. As the force decreases in the order $O(R^{-2})$ with $R$ being the distance between two objects, numerical solutions often omit the influence of objects that are far away. Analogously, in $E_{\mathrm{net}}$, we restrict the energy function to involve the interactions of \emph{neighboring nodes}. More specifically, we let $E_{\mathrm{net}}$ be a GNN with a scalar readout, i.e.,
\begin{align}\label{eq:Enet}
    E_{\mathrm{net}}: T^*Q_{\calG}\to \R.
\end{align}
According to \cref{thm:constantE}, if our system updates through \cref{eq:graph_obj}, $E_{\mathrm{net}}(q(t),p(t))$ remains constant. In other words, the principle law of energy conservation is obeyed in our model, while the exact formulation of the energy and the feature updating law is learnable to assist specific downstream tasks like node classification or link prediction.

\begin{figure*}[t!]
\centering
\includegraphics[width=0.8\textwidth]{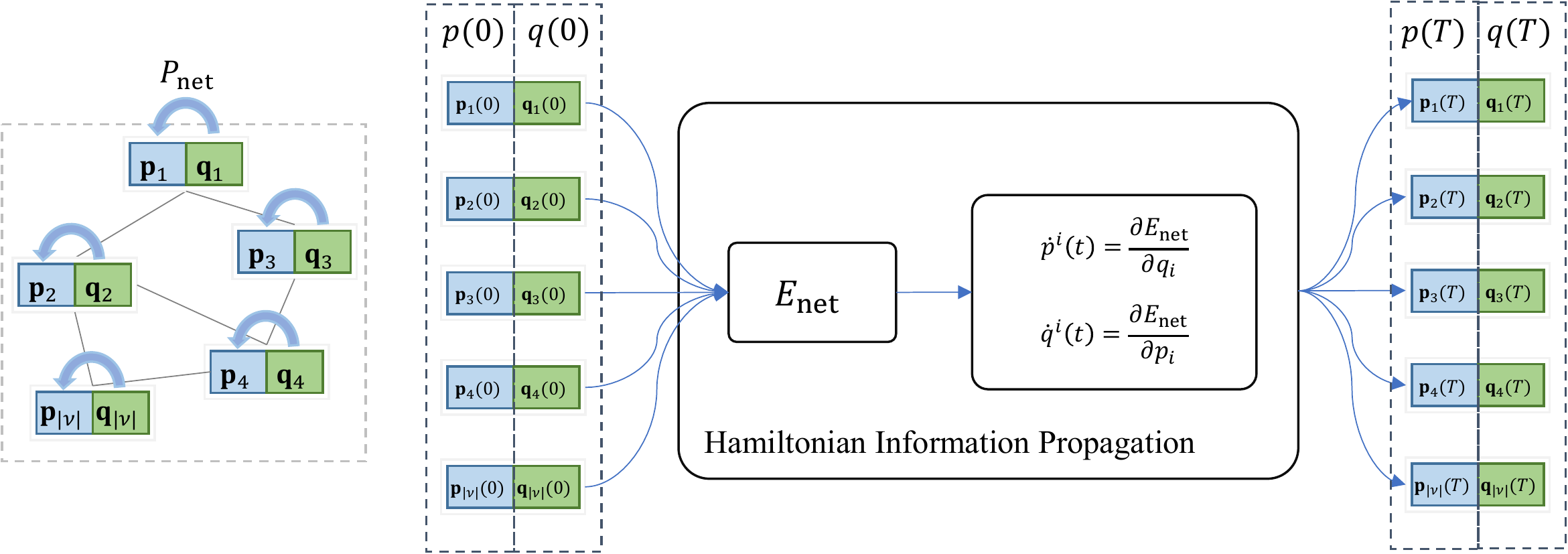}
\caption{The model architecture: each node  is assigned a learnable  ``momentum'' vector at time $t=0$ which will initialize the evolution of the system. The graph features evolve on a manifold following a learnable node embedding law \cref{eq:graph_obj} derived from the $E_{\mathrm{net}}$ analogous to the $n$-body evolves in the $\Real^3$ space following the basic physical laws \cref{eq:nbody}. At the time $t=T$, the canonical projection map \cref{appeq:co_pro} is performed to get the final node embedding. $E_{\mathrm{net}}(q(t),p(t))$ which takes the graph topology into consideration indicate a learnable graph energy function analogous to the total energy \cref{eq:total_E} of the $n$-body system.}
\label{fig:model_arch}
\end{figure*}

At time $t=T$, the solution of \cref{eq:graph_obj} is denoted as $(q(T),p(T))$ and the canonical projection map \cref{appeq:co_pro} is performed to obtain the final node embedding $q(T)$:
\begin{align*}
    \pi ((q(T),p(T))) = q(T).
\end{align*}
The full model architecture is illustrated in \cref{fig:model_arch} and summarized in \cref{alg:hamgnn}. 

\begin{algorithm}
\caption{HDG Node Embedding}\label{alg:hamgnn}
\begin{algorithmic}[1]
 \STATE \tb{Input: } $\calG = (\calV, \calE)$
 \STATE \tb{Output: }  Node embeddings $Z \in \mathbb{R}^{|\mathcal{V}| \times d}$.
 \STATE  Apply FC to compress raw node features and obtain $\{\bq_k\}_{k\in\calV}$.
\STATE  Apply $P_{\mathrm{net}}$ to obtain learnable momentum vectors $\{\bp_k\}_{k\in\calV}$.
\STATE Concatenate $\{\bq_k\}_{k\in\calV}$  to obtain $q$ according to \cref{eq.con_q}; concatenate $\{\bp_k\}_{k\in\calV}$  to obtain $p$ according to \cref{eq.con_p}. 
 \STATE  Perform Hamiltonian information propagation through \cref{eq:graph_obj} with the initial features at time $t=0$ being the input node features and learnable momentum, $(q(0), p(0)) = (q,p)$.
 \STATE Decouple $q(T)$ to $\{\bq_1(T),\ldots, \bq_{|\calV|}(T)\}$.
 \STATE Return node embeddings $Z$ with the $k$-th column being $\bq_k(T)$.
\end{algorithmic}
\end{algorithm}

\subsection{Discussions}\label{sec.mor_dis}
\begin{enumerate}
\item Comparison between HDG and the $n$-body system:
While our approach is inspired by Hamiltonian mechanics, we are not simulating an $n$-body system or solving a real-world physics problem. Our focus is on developing a new graph node embedding strategy where the propagation of information between nodes in a graph is inspired by Hamiltonian mechanics. In contrast to an $n$-body system, where all bodies interact directly through gravitational fields, HDG only allows interactions between a node and its local neighborhood. A learnable node embedding law is derived through gradient-based backpropagation for graph node embedding to aid downstream tasks such as node classification or link prediction. 

\item Stability of HDG: 
The $n$-body system is a well-known example of a chaotic system under most situations: a tiny change
in the initial state could result in significantly different evolutionary trajectories. However, in some special cases, such as the solar system, certain binary star systems, and some artificial systems where the mutual interactions between the bodies are controlled and small, the $n$-body system is not chaotic. One natural question is whether HDG, which follows a similar Hamiltonian principle, is chaotic or not. We argue that it is not chaotic for two reasons. Firstly, one can visually demonstrate convergence over time using t-SNE \cite{van2008visualizing} as shown in \cref{fig:converge}. We observe that the system is stable over time, with a small difference between the solution at $t=2$ and $t=5$.
\begin{figure}[h]
\centering
\subfloat[$t=1$]{\includegraphics[width=0.15\textwidth]{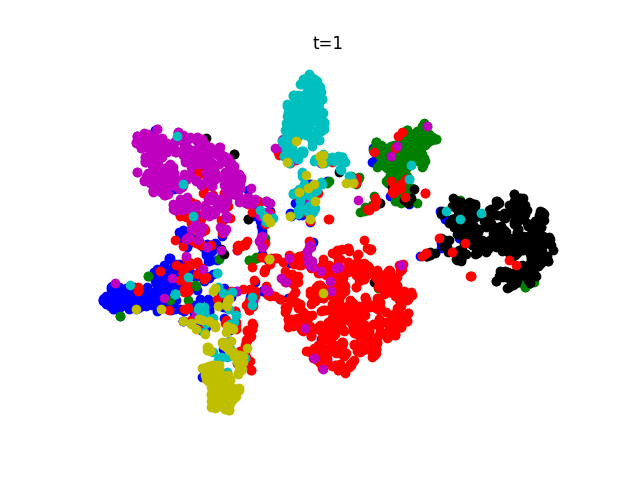} }
\hfil
\subfloat[$t=2$]{\includegraphics[width=0.15\textwidth]{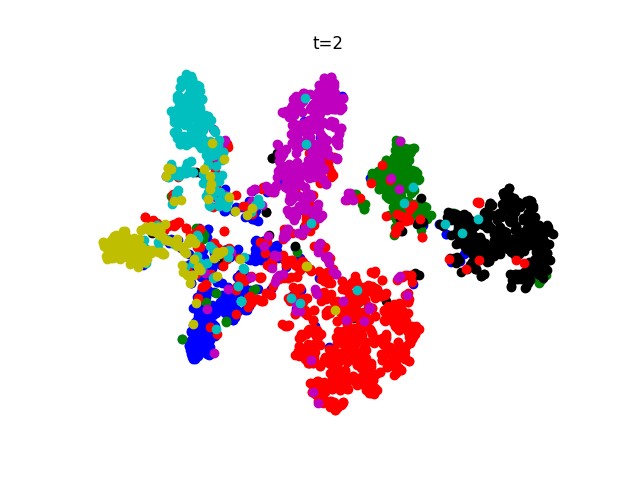}}
\hfil
\subfloat[$t=5$]{\includegraphics[width=0.15\textwidth]{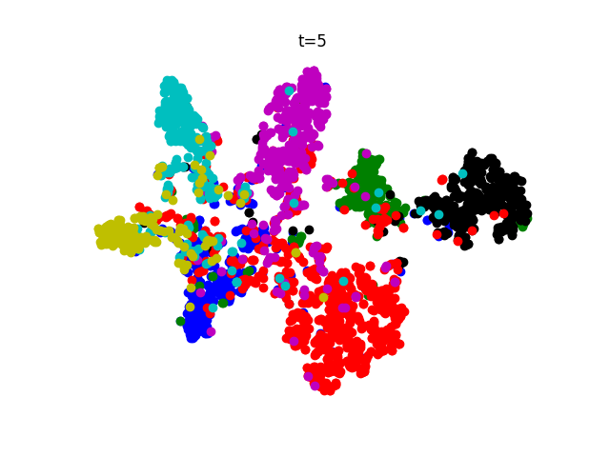}}
\caption{T-sne visualization of the node feature representation from HDG at different integration time $t$. From left to right: $t=1$, $t=2$ and $t=5$.}
\label{fig:converge}
\end{figure}
Secondly, we evaluate the performance of HDG against small perturbations, specifically adversarial attacks, to show that even with tiny changes in the initial state, the evolutionary trajectories do not change significantly. For the experimental evidence, we refer readers to \cref{sec:non-chaotic}.

\item Comparison between HDG and Riemannian manifold GNNs: 
In Riemannian manifold GNNs \cite{gu2018learning,bachmann2020constant_curvature,lou2020differentiating}, the energy function $E$ is typically defined as \cref{eq:gpq}, and the Hamiltonian equations \cref{eq:nbody} lead to the exponential map used for embedding graph nodes on some manifolds. HDG can therefore be seen as a generalization of Riemannian graph node embedding. However, there are also differences between HDG and Riemannian manifold GNNs.

One significant difference is that in HDG, the feature updating function and the message-passing function are combined into a single ODE function, whereas Riemannian manifold GNNs treat the node embedding process and node aggregation process as two distinct stages. Furthermore, in HDG, the energy function $E_{\mathrm{net}}$ is a learnable parameter optimized during training, while in Riemannian manifold GNNs, the energy function $E$ is usually predefined based on the manifold's geometry metric $g$. As shown in \cref{fig:hyper}, the Gromov $\delta$-hyperbolicity distributions\footnote{If the $\delta$-hyperbolicity distribution is more concentrated at lower values, the more hyperbolic the graph dataset.} \cite{Gromov1987} of various datasets exhibit a range of diverse values. This indicates that it is not optimal to embed a dataset with diverse geometries into a single globally homogeneous geometry.
The flexibility of HDG allows it to automatically learn the underlying geometry of any graph dataset without extensive tuning, even if it has diverse geometries. Experimental demonstration is provided in \cref{sec:mix}.
\end{enumerate}

\section{Experiments}\label{sec:exp}
In this section, we present a comprehensive evaluation of HDG, which includes comparisons to state-of-the-art Riemannian manifold GNNs, stability analysis, oversmoothness performance, and ablation studies. Specifically, we apply the HDG model, as described in Algorithm \ref{alg:hamgnn}, to learn node embeddings for datasets with various geometries, demonstrating its ability to automatically learn the underlying graph dataset geometry. Following the experimental settings of \cite{chami2019hyperbolicGCNN,gu2018learning,bachmann2020constant_curvature,lou2020differentiating,xiong2022pseudo}, we evaluate the effectiveness of the learned node embeddings by performing two downstream tasks: node classification and link prediction. To demonstrate that our learnable HDG as a Hamiltonian system is not chaotic, we test the model stability against adversarial attacks that serve as small perturbations of the input. Moreover, we investigate the effectiveness of HDG in mitigating the over-smoothing problem when many layers are stacked. Finally, we conduct ablation studies to demonstrate the efficiency of our model and the influence of ODE solvers on the model performance.

\subsection{Datasets and Baselines}
\begin{figure}[!tb]
    \centering
\includegraphics[width=0.45\textwidth,trim={1cm 0 1.6cm 1cm},clip]{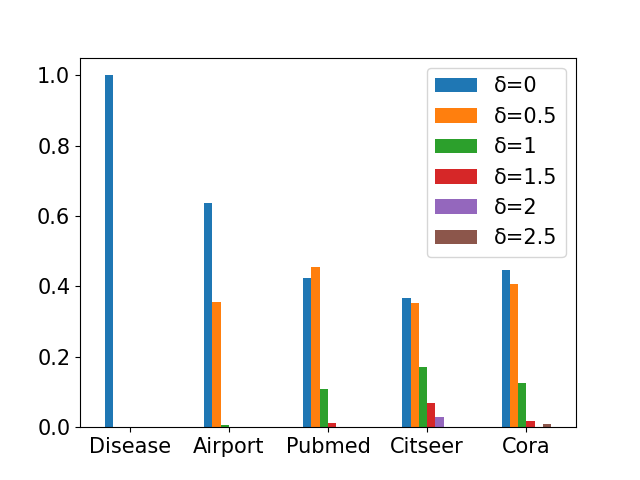}
    \caption{Gromov $\delta$-hyperbolicity distribution of datasets.}
    \label{fig:hyper}
\end{figure}

In this work, following \cite{chami2019hyperbolicGCNN,gu2018learning,bachmann2020constant_curvature,lou2020differentiating,xiong2022pseudo}, we select \tb{datasets with different geometries} to evaluate our approach for graph node embedding.
Specifically, we use three citation networks: Cora \cite{McCallum2004AutomatingTC}, Citeseer \cite{SenNamata2008}, Pubmed \cite{namata:mlg12-wkshp}; and two low hyperbolicity datasets \cite{chami2019hyperbolic_GCNN}: Disease and Airport. The statistical properties of these datasets are summarized in \cref{fig:hyper} and \cref{tab:bechmark}.
The citation datasets are widely used in graph representation learning tasks, and we adopt the same data splitting as previous work \cite{kipf2016semi}, randomly selecting 20 samples per class for training, 500 samples for validation, and 1000 samples for testing in each dataset. The low-hyperbolicity datasets Disease and Airport were proposed in \cite{chami2019hyperbolic_GCNN}, where Euclidean graph neural network models are ineffective at learning node embeddings. To ensure consistency, we follow the same data splitting and pre-processing steps as in \cite{chami2019hyperbolic_GCNN}.

Following the experimental settings of previous works \cite{chami2019hyperbolic_GCNN,gu2018learning,bachmann2020constant_curvature,lou2020differentiating,xiong2022pseudo}, we evaluate the effectiveness of the node embedding by performing \emph{two downstream tasks: node classification and link prediction} using the embeddings.
Rather than beating all the existing GNNs on these two specific tasks, we want to demonstrate that our flexible node embedding strategy is able to automatically learn, without extensive tuning, the underlying geometry of any given graph dataset even if its underlining geometry is very complex, e.g., a mixture of multiple geometries.  
To achieve this, we further create \tb{new datasets with more complex geometry} by combining Disease/Airport and Cora/Citeseer. These new datasets exhibit a mixture of hyperbolic and Euclidean local geometries (see \cref{sec:mix}).

\begin{table}[ht]
\centering
\begin{tabular}{lcccc}
\toprule
 Dataset & Nodes & Edges & Classes & Node Features   \\
 \midrule
 Disease &   1044    &   1043    &    2      &   1000     \\

 Airport &   3188    &  18631     &     4     &     4   \\
 
 Cora &     2708  &    5429   &     7     &    1433    \\

 Citeseer &    3327   &   4732    &    6      &   3703     \\

 Pubmed &  19717     &  44338     &      3   &  500      \\
 \bottomrule
\end{tabular}
\caption{Bechmark's statistics, where the dataset hyperbolicity is shown in \cref{fig:hyper}.}
\label{tab:bechmark}
\end{table}

In order to compare the performance of our proposed HDG approach to state-of-the-art Riemannian manifold graph neural networks (GNNs) in a fair manner, for the node classification task, we select several popular GNN models as the baseline, including \emph{the Euclidean GNNs}: GCN \cite{kipf2016semi}, GAT \cite{velickovic2018graph}, SAGE \cite{hamilton2017inductive}, and SGC \cite{wu2019simplifying}; \emph{the hyperbolic GNNs} \cite{chami2019hyperbolic_GCNN, liu2019hyperbolic_GNN}: HGNN, HGCN HGAT, and LGCN \cite{zhang2021lorentzian};  \emph{GNN that mixes Euclidean and hyperbolic embeddings}: GIL \cite{zhu2020GIL}; \emph{(Pseudo-)Riemannian GNNs}: $\kappa$-GCN \cite{bachmann2020constant_curvature} and 
$\calQ$-GCN \cite{xiong2022pseudo}; \emph{the graph neural diffusions}: GRAND \cite{chamberlain2021grand} and GraphCON \cite{rusch2022icml}. 
In addition, we include an MLP baseline that does not utilize graph topology information.
For the link prediction task, we compare HDG to the standard graph node embedding models including all the aforementioned baselines in the node classification task, except for the graph neural diffusion baselines, which are not considered node embedding models without open source code for link prediction.

We conduct the grid hyperparameter search for each dataset and report the test results of the best hyperparameters based on the validation accuracy. The specific hyperparameters can be found in \cref{tab:hyperpara} in the Appendix.
For HDG, we parameterize the energy function $E_{\mathrm{net}}$ in \cref{eq:Enet} using a two-layer GCN with \emph{tanh} activation function. The scalar output in \cref{eq:Enet} is calculated by the Frobenius norm of the GCN output.
We set $P_{\mathrm{net}}$  in \cref{eq:mom2} as a one-layer FC network that gives rise to a learnable momentum vector for each node. We use the numerical ODE solver \cite{torchdiffeq} for the Hamiltonian information propagation \cref{eq:graph_obj}.

\begin{table*}[!htb]
\centering
\resizebox{\textwidth}{!}{\begin{tabular}{ccccccccccc}
\hline\noalign{\smallskip}
Dataset  & \multicolumn{2}{c}{Disease} & \multicolumn{2}{c}{Airport} & \multicolumn{2}{c}{Pubmed} & \multicolumn{2}{c}{CiteSeer} & \multicolumn{2}{c}{Cora}\\
$\delta$-hyperbolicity & \multicolumn{2}{c}{0.0} & \multicolumn{2}{c}{1.0} & \multicolumn{2}{c}{3.5} & \multicolumn{2}{c}{4.5} & \multicolumn{2}{c}{11.0}\\
\hline\noalign{\smallskip}
Method & LP & NC & LP & NC & LP & NC & LP & NC & LP & NC \\
\hline\noalign{\smallskip}
MLP &  \third{83.37$\pm$5.04}  &50.00$\pm$0.00   &  87.04$\pm$0.56  &76.96$\pm$1.77 &  88.69$\pm$1.59  &71.95$\pm$1.38 &  89.65$\pm$1.00  &58.10$\pm$1.87  &  91.07$\pm$0.56 &57.15$\pm$1.15   \\
HNN &  81.37$\pm$8.78  &56.40$\pm$6.32 &  86.06$\pm$2.08  &80.49$\pm$1.54  &  94.69$\pm$0.25  &71.60$\pm$0.47 &  89.83$\pm$0.39  &55.13$\pm$2.04  &  92.83$\pm$0.76 &58.03$\pm$0.55   \\
\hline
GCN &  60.38$\pm$2.51  &81.10$\pm$1.33  &  90.97$\pm$0.65  &82.25$\pm$0.56  &  91.37$\pm$0.09  &77.83$\pm$0.77  &  93.20$\pm$0.28  &\third{71.78$\pm$0.34} &  92.89$\pm$0.77 & 80.29$\pm$2.29 \\
GAT &  62.03$\pm$1.58  &70.40$\pm$0.49 &  91.05$\pm$0.83  &81.59$\pm$0.36  &  91.03$\pm$0.67  &{78.21$\pm$0.44} &  93.83$\pm$0.65  &71.58$\pm$0.80  &  93.34$\pm$0.50 &\third{83.03$\pm$0.50} \\
SAGE &  68.02$\pm$0.43  &81.60$\pm$7.68 &  91.40$\pm$0.88  &81.97$\pm$0.85  &  93.61$\pm$0.26  &77.63$\pm$0.15  &  93.37$\pm$0.88  & 70.60$\pm$0.47 &  92.94$\pm$0.40 & 81.80$\pm$0.65 \\
SGC &  59.83$\pm$4.01  &82.78$\pm$0.93  &  89.72$\pm$0.82  &81.40$\pm$2.21 &  92.16$\pm$0.13  &76.83$\pm$1.11  &  94.78$\pm$0.77  &70.88$\pm$1.32 &  93.15$\pm$0.22 &81.98$\pm$1.71 \\
\hline
HGNN &  81.54$\pm$1.22  &80.51$\pm$5.70 &  92.46$\pm$0.20  &84.54$\pm$0.72 &  93.09$\pm$0.09  &76.65$\pm$1.38 &  90.35$\pm$0.57  & 69.43$\pm$0.99 &  92.05$\pm$0.33 & 79.53$\pm$0.98 \\
HGCN &  90.80$\pm$0.31  &89.87$\pm$1.13  &  94.28$\pm$0.20  &85.35$\pm$0.65 &  \third{96.79$\pm$0.01}  &76.38$\pm$0.81  &  93.60$\pm$0.14  &67.60$\pm$0.93  &  94.10$\pm$0.05 &78.70$\pm$0.96  \\
HGAT &  87.63$\pm$1.67  &\third{90.30$\pm$0.62} &  94.64$\pm$0.51  &89.62$\pm$1.03  &  96.86$\pm$0.03  &77.42$\pm$0.66  &  93.45$\pm$0.25  &68.64$\pm$0.30 &  94.96$\pm$0.36 &78.32$\pm$1.39   \\

LGCN &  96.80$\pm$0.40  & 88.47$\pm$1.80 &  96.98$\pm$0.54  & 88.22$\pm$0.18  &  \second{96.90$\pm$0.00}  &77.35$\pm$1.38  &  96.40$\pm$0.20  & 68.08$\pm$1.98 &  94.40$\pm$0.20 & 80.60$\pm$0.92   \\

\hline
\exc{GIL}  &  \first{99.97$\pm$0.08}  &\second{90.70$\pm$0.40} &   \second{97.92$\pm$2.64} &\second{91.33$\pm$0.78} &  \exc{91.22$\pm$3.25}  &\exc{78.91$\pm$0.26} &  \third{95.99$\pm$8.89}  &\exc{72.97$\pm$0.67} &  \second{97.78$\pm$2.31} &\second{83.64$\pm$0.43}  \\
\hline
$\kappa$-GCN  &- & -   &\third{96.35$\pm$0.62} &87.92$\pm$1.33    &96.60$\pm$0.32 &\third{79.20$\pm$0.65}   &95.79$\pm$0.24&\second{73.25$\pm$0.51}   &94.04$\pm$0.34&81.08$\pm$1.45\\
$\calQ$-GCN &-  & 70.79$\pm$1.23  &96.30$\pm$0.22 &\third{89.72$\pm$0.52}  &{96.86$\pm$0.37} &\first{81.34$\pm$1.54}  &\second{97.01$\pm$0.30} &\first{74.13$\pm$1.41}  &\third{95.16$\pm$1.25} &\first{83.72$\pm$0.43} \\
\hline
HDG    &  \second{97.64$\pm$1.88}    & \first{91.27$\pm$5.94}        &  \first{99.99$\pm$0.01}    & \first{96.57$\pm$0.76}           &  \first{97.79$\pm$3.13}                   & \second{79.47$\pm$0.45}         & \first{99.22$\pm$0.13}   & 70.76$\pm$0.98    &  \first{99.13$\pm$1.32}  &  82.43$\pm$0.49   \\
\bottomrule
\end{tabular}}
\caption{Node classification (NC) accuracy and link prediction (LP) ROC. The best and second-best for each criterion are highlighted in \first{red} and \second{blue}, respectively. ``-'' indicates the open source code and/or the result is not available. 
}\label{tab:lp-nc} 
\end{table*}

\subsection{Node Embedding Performance}\label{sec.nod_emb}
\begin{table}[!htb]\small
    \centering
    \setlength{\tabcolsep}{1pt}
 \resizebox{0.5\textwidth}{!}{\begin{tabular}{c|c|c|c|c|c}
    \hline
        Method & Disease & Airport & Pubmed & Citeseer & Cora\\
        \hline
         GRAND   & 74.52$\pm$3.37 & 60.02$\pm$1.55  &  \second{79.32$\pm$0.51}  &\first{71.76$\pm$0.79} & \first{82.80$\pm$0.92}  \\
        GraphCON & \second{87.50$\pm$4.06} & 68.61$\pm$2.10  & 78.87$\pm$0.97  & \second{71.33$\pm$0.83} & \second{82.49$\pm$1.08}   \\
        HDG   & \first{91.27$\pm$5.94}  &\first{96.57$\pm$0.76}    &\first{79.47$\pm$0.45}  &70.76$\pm$0.98   &  {82.43$\pm$0.49}    \\
         \hline
    \end{tabular}}
    \caption{Comparison between HDG and graph diffusion models on the node classification task. We set the constant ODE block and Laplacian ODE function in GRAND with no rewiring.}
    \label{tab:grand}
\end{table}

\tb{Node Classification:}
The node classification performance on the benchmark datasets using the baseline models and the proposed HDG is shown in \cref{tab:lp-nc,tab:grand}. We observe that HDG achieves good performance on all the datasets.
More specifically, HDG achieves the highest node classification accuracy on the two datasets with the lowest $\delta$-hyperbolicity, i.e., Disease and Airport. 
The HDG model improves the absolute accuracy by a large margin of 5.24\% on the Airport dataset as compared with the GIL model, which is considered to be a strong competitor. 
For datasets with high $\delta$-hyperbolicity, HDG performs comparably to Euclidean-based GNNs and outperforms all hyperbolic GNNs. This is because hyperbolic GNNs are designed for datasets with lower $\delta$-hyperbolicity and struggle to adapt to those with higher values. In contrast, similar to $\calQ$-GCN and $\kappa$-GCN, which consider mixed geometry, HDG is capable of automatically learning the underlying geometry, resulting in consistently good performance across different structured datasets. 
Compared to the graph neural diffusion models presented in \cref{tab:grand}, we observe that our Hamiltonian information propagation mechanism has a significant advantage, particularly on the Disease and Airport datasets where the baseline diffusion models struggle to adapt.



We would like to emphasize that HDG, which utilizes a GCN as $E_{\mathrm{net}}$ with new Hamiltonian information propagation \cref{eq:graph_obj}, significantly outperforms vanilla GCN with low-pass convolutional information propagation on hyperbolic datasets such as Disease (from 81.10\% to 91.27\%) and Airport (from 82.25\% to 96.57\%). This remarkable improvement demonstrates the advantages of the Hamiltonian information propagation mechanism used in our model.

\tb{Link Prediction:} To further demonstrate that HDG can automatically learn the underlying geometry, we include the link prediction task as another node embedding downstream task as shown in \cref{tab:lp-nc}. For the link prediction task, we use negative sampling to generate the fake edges in the graph and we report the ROC-AUC score in \cref{tab:lp-nc}. We follow the data split in \cite{chami2019hyperbolicGCNN,zhu2020GIL}. To calculate the probability score between pairs of edges, we use the Fermi-Dirac decoder in \cite{nickel2017poincare} for link prediction tasks.
We observe that for this task, HDG still adapts well to all datasets and achieves the best performance on all datasets except the Disease dataset. 
These findings confirmed the advantage of HDG in adapting to diverse graph datasets.


In conclusion, HDG, which leverages a learnable $E_{\mathrm{net}}$ to guide Hamiltonian information propagation as defined in \cref{eq:graph_obj}, exhibits the flexibility to learn the underlying geometry of any graph dataset without the need for extensive tuning, even when the geometry varies from one dataset to another as shown in \cref{fig:hyper}. On the other hand, Riemannian manifold GNNs that use fixed energy functions $E$ as defined in \cref{eq:gpq} do not adapt as effectively as HDG.



\subsection{Mixed Geometry Datasets}\label{sec:mix}
To further demonstrate the versatility of our node embedding strategy in adapting to diverse geometries, including cases where the geometry varies within a single graph dataset, we create new mixed-geometry graph datasets by combining the datasets used in \cref{sec.nod_emb}.

\subsubsection{Mixed Airport + Cora Dataset} 
We start by combining the Airport (3188 nodes) and Cora (2708 nodes) datasets into a single large graph with 5896 nodes. The resulting graph contains a mixture of hyperbolic and Euclidean geometries, with the Airport dataset exhibiting a more hyperbolic geometry and the Cora dataset exhibiting a more Euclidean geometry, as shown in Figure \ref{fig:hyper}. We selected these two datasets because they have a similar number of nodes.
To unify the node feature dimension, we pad zero elements in the input feature vector of each dataset.
We did not introduce any new edges in the new dataset. Instead, nodes from the Airport dataset are connected to their neighbors from the same dataset, and similarly for the nodes from the Cora dataset. The resulting adjacency matrix is a 2-by-2 block matrix, where the diagonal blocks are the adjacency matrices of the Airport and Cora datasets, and the off-diagonal blocks are full zero matrices. We split the data into training, validation, and test sets using a random split of 60\%, 20\%, and 20\%, respectively.

\subsubsection{Other Mixed Datasets} 
We used the same methodology as for the mixed Airport + Cora dataset to create three additional mixed datasets: mixed Airport + Citeseer, mixed Disease + Cora, and mixed Disease + Citeseer. The resulting datasets have different geometries in each component compared to their corresponding base datasets.

We evaluate HDG against several baselines, including GCN, HGCN, GIL, and $\calQ$-GCN, on the mixed-geometry graph datasets, and present the results in \cref{tab:mix_geo}. We find that the baselines are not able to adapt effectively to the diverse geometries present in the data. In contrast, HDG, which uses novel Hamiltonian information propagation, performs the best on three mixed datasets, further demonstrating that our learnable node embedding law is effective in guiding the learning of the underlying geometry, as discussed in \cref{sec.mor_dis}. HDG achieves 95.63\% on the mixed Airport + Cora dataset, surpassing the second-best $\calQ$-GCN model by a significant margin of 5.48\%. On the mixed Disease + Cora dataset, HDG achieves the second-best performance, which is only 0.88\% lower than $\calQ$-GCN.

\begin{table}[h]\small
    \centering
 \resizebox{0.5\textwidth}{!}{\setlength{\tabcolsep}{1pt}
 \begin{tabular}{c|c|c|c|c|cc}
    \toprule
        datasets & GCN & HGCN & GIL & $\calQ$-GCN  & HDG \\
        \hline
        Air+Cora & 75.16$\pm$0.65 & 78.72$\pm$0.42  & 82.04$\pm$1.27  & 90.25$\pm$0.81 & \tb{95.63$\pm$0.46} \\
          \hline
               Air+Cite & 70.97$\pm$0.34 & 74.65$\pm$0.40  &  77.83$\pm$1.57  & 87.63$\pm$0.28 & \tb{92.20$\pm$0.63} \\
         \hline
            Disease+Cora & 93.49$\pm$0.23  & 94.90$\pm$0.41  &  94.50$\pm$0.46  & \tb{95.88$\pm$0.12} & {95.00$\pm$0.59} \\
         \hline
            Disease+Cite & 87.17$\pm$0.44 & 88.61$\pm$0.30  &  88.24$\pm$0.47  & 88.24$\pm$0.27  & \tb{90.02$\pm$0.72} \\
      \bottomrule
    \end{tabular}}
    \caption{Node classification on the Mixed Geometry Dataset. First row: mixed Airport+Cora dataset; Second row: mixed Airport+Citeseer dataset; Third row: mixed Disease+Cora dataset; Forth row: mixed Disease+Citeseer dataset.
    }
    \label{tab:mix_geo}
\end{table}

\subsection{Stability of HDG}\label{sec:non-chaotic}
Both HDG and the physical $n$-body system are Hamiltonian systems, where the system evolves following the trajectory as solutions of the {Hamiltonian equation} (cf. \cref{eq:nbody} or \cref{eq:graph_obj}). As we have discussed in \cref{sec.mor_dis}, there are differences between HDG and the $n$-body system. While the $n$-body system is a chaotic system, we argue that HDG under the learnable node embedding law is a stable system. We have already presented evidence of convergence in \cref{fig:converge}, and we now show that the performance of HDG against small perturbations, specifically adversarial attacks, does not change significantly.

To test the stability of HDG, we apply the two strongest injection attacks reported in the Graph Robustness Benchmark (GRB) \cite{zheng2021grb}: SPEIT \cite{qin2020speit} and TDGIA \cite{zou2021tdgia}. We also implement the Nettack poisoning evasion attack \cite{zugner2018adversarial} for comparison. We evaluate the transductive graph learning settings on the Cora dataset with the maximum number of injected nodes and edges set to 50/50 for SPEIT and TDGIA attacks. We follow the Nettack-direct attack in \cite{zugner2018adversarial} that directly attacks 40 selected target nodes. The reported accuracy under the Nettack-direct attack is calculated based on the target nodes. From the results in \cref{tab.cora_att}, we observe that HDG achieves a better node classification result than GAT and GRAND even when subject to various adversarial attacks. This shows that, unlike most physical $n$-body systems, HDG under the learnable node embedding law is not a chaotic system, as it remains stable against small perturbations. The reason for this stability may come from the differences between HDG and the $n$-body system, as discussed in \cref{sec.mor_dis}. Further theoretical understanding of this stability is left for future work.

\begin{table}[h]
\centering
\begin{tabular}{ccccccc}
\toprule
Attack    & HDG & GAT & GRAND &  \\
\midrule
SPEIT & \tb{79.66}    & 39.16       & 56.58         &  \\
TDGIA & \tb{75.08}    & 50.97       & 52.86         &  \\
Nettack & \tb{63.50}    & 37.75       & 54.50         & \\
\bottomrule
\end{tabular} 
 \caption{Node classification accuracy on the Cora Dataset with adversarial perturbations.}\label{tab.cora_att}
\end{table}
\subsection{Resilience to Over-Smoothing}\label{sec:ove_smo}

\begin{table*}[!t]
\centering
\makebox[\textwidth][c]{
\begin{tabular}{lcccccccc}
\toprule
Dataset  &  Models  &  2 layers   &  4 layers   &  8 layers   &  16 layers  &  32 layers  &  64 layers   \\
\midrule
\multirow{3}{*}{Pubmed} 

& GCN &  79.20$\pm$0.23  &  77.30$\pm$1.45   &  66.20$\pm$13.46   &   42.48$\pm$2.49   &  41.74$\pm$2.82   &  40.70$\pm$0.0   \\
& HGCN &  75.82$\pm$0.22  &  75.56$\pm$1.67   &  59.60$\pm$4.62   &   42.62$\pm$ 3.63  &  40.70$\pm$0.0   &  39.44$\pm$0.92\\
& HDG &  78.52$\pm$0.49  &  78.37$\pm$0.71  &  78.87$\pm$0.99  &   79.45$\pm$0.53  &   78.05$\pm$1.33  &   77.27$\pm$2.76   \\
\midrule
\multirow{3}{*}{Citeseer} 

& GCN &  72.62$\pm$0.51  &  60.42$\pm$2.49   &  31.24$\pm$12.98   &   22.02$\pm$2.15   &  22.20$\pm$1.59   &  23.24$\pm$1.06    \\
& HGCN &  67.60$\pm$0.93  &  50.54$\pm$2.29   &  18.10$\pm$0.0   &   17.82$\pm$0.52   &  19.10$\pm$2.24   &  17.88$\pm$0.55   \\
& HDG &  70.39$\pm$1.04  &  70.03$\pm$0.80  &  70.25$\pm$1.10  &   69.96$\pm$1.08  &   69.68$\pm$0.67  &   68.94$\pm$0.88 \\
\midrule
\multirow{3}{*}{Cora} 
& GCN &  81.94$\pm$0.30  &  75.60$\pm$3.01  &  34.76$\pm$11.03  &   27.14$\pm$4.34  &   23.58$\pm$6.57  &  27.72$\pm$2.81   \\

& HGCN &  78.62$\pm$2.03  &  59.76$\pm$6.99  &  34.72$\pm$4.29  &   31.90$\pm$0.00  &   31.90$\pm$0.00  & 24.62$\pm$9.98  \\
& HDG &  80.96$\pm$0.76  &  82.00$\pm$0.92  &  82.25$\pm$0.85  &   82.24$\pm$0.83  &   81.52$\pm$0.81  &   80.05$\pm$1.30   \\

\bottomrule
\end{tabular}}
\caption{Node classification accuracy(\%) when increasing the number of layers on the citation datasets.}
\label{tab:num_layer_full}
\end{table*}

\begin{figure}[!htb]
\centering
\includegraphics[width=0.4\textwidth]{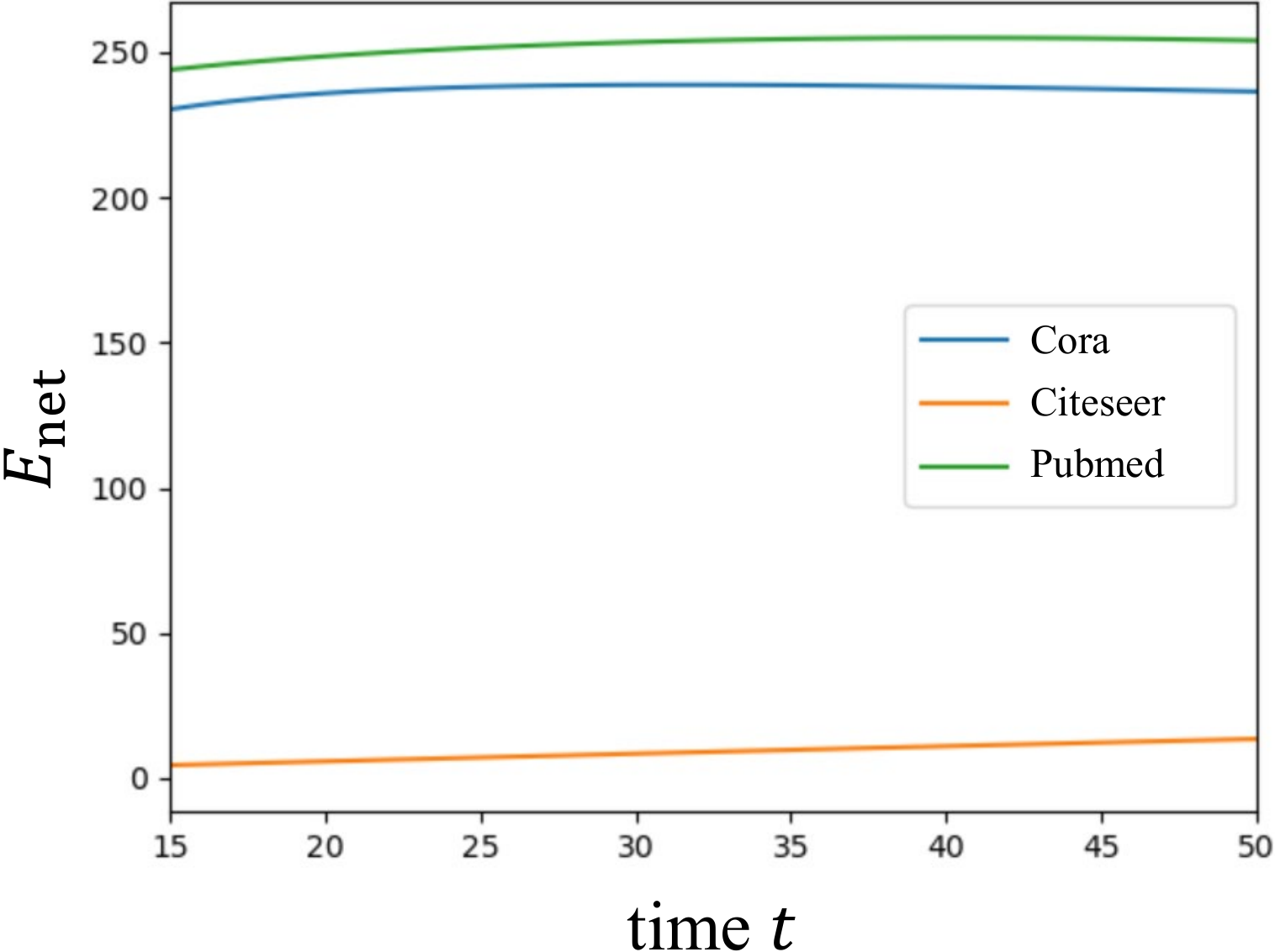}
\caption{Visualization of the Hamiltonian energy function $E_{\mathrm{net}}$'s scalar output over time.}
\label{fig:energy_plot}
\end{figure}

To demonstrate HDG's ability to mitigate over-smoothing problems, we carried out an experiment using the Euler fixed step-size ODE solver \cite{chen2018neural} to show its good performance with many GNN layers. The ODE system using a fixed step-size solver with a step size $\tau =1.0$ and an integer integration time $T$ can be viewed as stacking up $T$ GNN layers.
As shown in \cref{tab:num_layer_full}, when we stack more HDG layers by setting  larger time $T$, the accuracy of their node classification ability does not drop as sharply as it does for other types of GNNs including GCN and HGCN.
One reason for this stability against over-smoothing is that the node feature energy is fixed during the feature updating along the Hamiltonian flow. We additionally plot the HDG's Hamiltonian energy function $E_{\mathrm{net}}$ output over time 
in \cref{fig:energy_plot}. We observe the energy $E_{\mathrm{net}}$ over time nearly keeps constant, which is expected as \cref{thm:constantE} has indicated this fact. For different datasets, the final learned energy of the full graph may vary a lot. We leave it for future work about how to link the learnable energy to the graph structures. Note, during the feature updating, the energy $E_{\mathrm{net}}$ scalar output is not directly utilized. Instead, only the derivative of $E_{\mathrm{net}}$ in \cref{eq:graph_obj} guides the graph feature updating through a learnable law analogous to the $n$-body system's basic physical laws.

\subsection{Ablation Studies}
\subsubsection{Vanilla ODE vs. HDG}
We compare HDG with a vanilla neural ODE \cref{eqn:vanODE}:
\begin{align}\label{eqn:vanODE}
    \dot{q}(t) = f_{\mathrm{net}}(q(t)), 
\end{align}
where $f_{\mathrm{net}}$ is composed of two FC layers and a non-linear activation function. 
\begin{table}[!htp]
\centering
\begin{tabular}{c c c c}
\toprule
Dataset  & Pubmed & Citeseer  & Cora     \\
\midrule
Vanilla ODE & 73.30$\pm$3.31  &  56.60$\pm$1.26  &  68.38$\pm$1.18 \\
HDG & {\bf 79.47$\pm$0.45}  &  {\bf 70.76$\pm$0.98}  &  {\bf 82.43$\pm$0.49} \\
\bottomrule
\end{tabular}
\caption{Comapre HDG with vanilla ODE without Hamiltonian information propagation.}
\label{tab:different_gnn}
\end{table}
The results are shown in \cref{tab:different_gnn} where we can see HDG outperforms the vanilla ODE by a big margin. This demonstrates the advantage of our Hamiltonian information propagation \cref{eq:graph_obj}.
\subsubsection{ODE solver}
To show the influence of ODE solvers, we compare the different ODE solvers in \cite{chen2018neural}, including the fixed step size solver Euler and Rk4, and adaptive step size solver Dopri5. The results are reported in \cref{tab:different_odesolver}, and we can see that the performances of HDG using different ODE solvers are very close. In other words, HDG performs equally well with different ODE solvers.  
\begin{table}[!htp]
\centering
\begin{tabular}{c c c c}
\toprule
Dataset  & Pubmed & Citeseer  & Cora     \\
\midrule
Euler & 79.47$\pm$0.45  &  70.76$\pm$0.98  &  82.43$\pm$0.49 \\
Rk4 &  79.06$\pm$0.68  &  70.56$\pm$1.12  &  82.13$\pm$0.62 \\
Dopri5 & 78.78$\pm$1.44  &  69.52$\pm$0.98  &  82.13$\pm$0.52 \\

\bottomrule
\end{tabular}
\caption{HDG using different ODE solver. }
\label{tab:different_odesolver}
\end{table}
One drawback of the ODE solvers provided in \cite{torchdiffeq} is that they are not guaranteed to have the energy-preserving property in solving the Hamiltonian equations. However, since we are neither simulating  an $n$-body system nor solving a real-world $n$-body physics problem, this flaw does not significantly deteriorate our model performance regarding the embedding adaptation to datasets with various structures. 
Our extensive experiments on the node classification and link prediction tasks have demonstrated that the solvers provided in \cite{torchdiffeq} are sufficient for our use. 
We leave the use of Hamiltonian equation solvers for future work to investigate whether solvers with the energy-preserving property can better help graph node embedding or mitigate the over-smoothing problem.

\subsubsection{Inference Time}\label{sec:time}
We compare the inference time of HDG with other GNNs, including GCN, HGCN, and GraphCON. As shown in Table \ref{tab:time}, HDG has an inference time of 5.76 ms, slightly higher than GraphCON and GCN but lower than HGCN. 
The additional step to compute the derivative and integration in the Hamiltonian information propagation, as shown in \cref{eq:graph_obj}, incurs the extra time taken for HDG compared to GCN. Nonetheless, the inference time of HDG is sufficiently low for real-time operations, making it a practical choice for the graph node embedding task.

\begin{table}[htp]\small
\centering
\begin{tabular}{ccccc}
\toprule
Method  & GCN & HGCN & GraphCON & HDG  \\
\midrule
Inference Time(ms) &   2.96 & 6.14 & 4.26  & 5.76 \\

\bottomrule
\end{tabular}
\caption{Inference time of models. GCN and HGCN are based on one hidden layer with a hidden dim of 64. GraphCON and HDG are based on Euler method with an integral time of 1. }
\label{tab:time}
\end{table}




\section{Conclusion}\label{sec:con}
In this paper, we introduce a novel Hamiltonian graph neural network approach called HDG, which utilizes Hamiltonian mechanics to propagate graph node information. Our approach combines the feature updating and message-passing functions into one ODE function and uses a learnable energy function to optimize the node embeddings. We demonstrate the effectiveness of HDG on various graph datasets with diverse geometries, verifying that it can automatically learn the underlying geometry without extensive tuning. Our model outperforms state-of-the-art Riemannian manifold GNNs on node classification and link prediction tasks. We conducted stability and over-smoothing analysis as well as ablation studies to validate the efficacy and efficiency of HDG. Our work sheds light on the potential of applying Hamiltonian mechanics to GNNs and opens up new avenues for future research in this area.

\bibliographystyle{IEEEtran}
\bibliography{aaai23,IEEEabrv}

\newpage

 




\appendix

The following \cref{tab:hyperpara} records the hyper-parameters used in our experiments.

\begin{table}[!htb]\small
\centering
\resizebox{\columnwidth}{!}{
\begin{tabular}{c|cccccc}
    \toprule
    Dataset & Task & Lr & Decay & Dropout  & Time & Step size  \\
    \midrule
    \multirow{2}{*}{Disease} & NC & 0.005 & 0.0001 & 0.4  & 1.0 & 1.0 \\
    & LP& 0.005 & 0.01 & 0.4  & 1.0 & 0.5\\

    \multirow{2}{*}{Airport} & NC & 0.005 & 0.001 & 0.8  & 2.0 &0.5 \\
    & LP & 0.005 & 0.0001 & 0.0  & 1.0 & 1.0\\

    \multirow{2}{*}{Pubmed} & NC & 0.005 & 0.01 & 0.2  & 8.0 & 1.0 \\
    & LP & 0.005 & 0.0001 & 0.0 & 1.0 & 0.5 \\

     \multirow{2}{*}{Citeseer} & NC & 0.01 & 0.001 & 0.2 & 1.0 & 0.5 \\
    & LP & 0.001 & 0.0001 & 0.0  & 10.0 & 1.0  \\

    \multirow{2}{*}{Cora} & NC & 0.005 & 0.1 & 0.6 & 10.0 & 1.0 \\
    & LP & 0.005 & 0.0001 & 0.4  & 1.0 & 1.0 \\
    \bottomrule
\end{tabular}}

\caption{Hyper-parameters used in \cref{tab:lp-nc} }
\label{tab:hyperpara}
    
\end{table}



\end{document}
\endinput